\documentclass[10pt,a4paper,twocolumn,USenglish]{article}

\usepackage[top=.7in,left=.7in,right=.7in,bottom=1.1in]{geometry}
\usepackage{setspace}

\usepackage{authblk}
\usepackage{fourier}
\usepackage[USenglish]{babel}
\usepackage[utf8]{inputenc}
\usepackage{authblk}
\usepackage{makecell}
\usepackage{multirow}
\usepackage{tabularx}
\usepackage{booktabs}
\usepackage[export]{adjustbox}
\usepackage{subfig}
\usepackage{url}
\usepackage[bookmarks, colorlinks, breaklinks]{hyperref}
\usepackage[usenames,dvipsnames,table]{xcolor}
\usepackage{colortbl}
\usepackage{setspace}
\usepackage{caption}
\usepackage{cite}
\usepackage[style=british]{csquotes}
\usepackage{outlines}
\usepackage[linewidth=0.5pt]{mdframed}
\usepackage{subfig}
\usepackage{booktabs}
\usepackage{rotating}
\usepackage{xspace}
\usepackage{graphicx}
\usepackage{lipsum}
\usepackage{enumitem}

\definecolor{LightCyan}{rgb}{0.88,1,1}
\definecolor{MyGray}{gray}{.5}

\title{
    {\huge\bfseries MajinBook}
    \vspace{0.4em}
    \\ 
    {An open catalogue of digitally mediated world literature}
}

\author[1,2]{Antoine Mazi\`eres\thanks{Corresponding author: mazr@ik.me}}
\author[1]{Thierry Poibeau}
\affil[1]{\href{https://www.lattice.cnrs.fr}{Lattice} (ENS-PSL, CNRS), Montrouge, France}
\affil[2]{\href{https://leda.dauphine.fr/}{LEDa} (Dauphine-PSL, CNRS), Paris, France}

\date{}

\begin{document}

\newcommand{\ffr}{\textsc{ffr}}
\newcommand{\tb}[1]{\textcolor{blue}{#1}}
\newcommand{\m}[1]{\tb{#1}}
\newcommand{\apr}[1]{\textcolor{Apricot}{#1}}
\newcommand{\todo}[1]{{\sc\apr{[#1]}}}
\hypersetup{linkcolor= MidnightBlue,citecolor= MidnightBlue,filecolor=black,urlcolor= MidnightBlue}
\definecolor{gray}{rgb}{0.98,.98,.98}
\definecolor{darkgray}{rgb}{0.8,.8,.8}
\definecolor{darkdarkgray}{rgb}{0.66,.66,.66}
\newcommand{\tg}[1]{\textcolor{green}{#1}}
\newcommand{\g}[1]{\tg{#1}}
\newcommand{\tr}[1]{\textcolor{red}{#1}}
\renewcommand{\r}[1]{\tr{#1}}

\maketitle

\begin{abstract}

This data paper introduces \emph{MajinBook}, an open catalogue designed to facilitate the use of shadow libraries---such as Library Genesis and Z-Library---for computational social science and cultural analytics. By linking metadata from these vast, crowd-sourced archives with structured bibliographic data from Goodreads, we create a high-precision corpus of over 539,000 references to digitally mediated English-language books. Spanning three centuries and reflecting a contemporary selection bias, these entries are enriched with first publication dates, genres, and popularity metrics like ratings and reviews. Our methodology prioritises natively digital \textsc{epub} files to ensure machine-readable quality, while addressing biases in traditional corpora like HathiTrust, and includes secondary datasets for French, German, and Spanish. We evaluate the linkage strategy for accuracy, release all underlying data openly, and discuss the project’s legal permissibility under \textsc{eu} and \textsc{us} frameworks for text and data mining in research.

\vspace{1em}
\vspace{1em}

\end{abstract}

\section*{Introduction}

The use of text collections as a basis for linguistic inquiry has a long history that far predates modern computing. However, it was the computational turn of the late 1950s that established this practice as a fully-fledged academic discipline, later known as corpus linguistics. Dependent on technology capable of managing large quantities of machine-readable text~\cite{McEnery2003}, early corpus linguistics often aimed to study the building blocks of language---e.g., grammar and lexis. With the increasing availability of computational power, other subfields soon broadened the practice’s aspirations. Sociolinguistics and discourse analysis are prominent illustrations of this methodological borrowing, applying corpus techniques to the study of social structures and norms. Similarly, the field of distant reading applies these methods to uncover patterns in literary history. In this context, the recent progression of Computational Social Science (\textsc{css}) can be seen as an advance in scale---utilising millions of books and social media data---and in tools---applying advanced machine learning and network analysis---rather than a fundamental change in the long-standing goal of understanding culture through texts.

The vision of using extensive book corpora to \emph{represent} culture at a large scale was significantly advanced by mass digitisation projects. The scanning effort undertaken by Google, beginning in 2002, culminated in the Google Books project and yielded, among other resources, the dataset for a foundational \enquote{quantitative analysis of culture}~\cite{michel2011quantitative}. Shortly thereafter, the HathiTrust Digital Library~\cite{christenson2011hathitrust} was formed as a partnership between Google and major research libraries. Now containing over 18 million volumes, HathiTrust has become a go-to resource for making broad claims about culture through the computational analysis of books. Its utility is demonstrated by its application across diverse fields, including the history of science~\cite{murdock2017multi}, literary history~\cite{underwood2019distant}, gender studies~\cite{underwood2018transformation}, and musicology~\cite{downie2020hathitrust}.

Despite its scale and utility, HathiTrust is not without significant limitations and biases, many of which stem from its origins. The collection itself is not a neutral representation of world literature; its composition privileges large research universities, while effectively excluding smaller, non-affiliated institutions and other forms of participation. Furthermore, as a collection built from scanned physical books, the quality of the underlying text can be inconsistent due to data integrity issues and errors from the Optical Character Recognition (\textsc{ocr}) process. The most significant challenge, however, relates to access. The \enquote{dark history}~\cite{centivany2017dark} of HathiTrust’s formation was fraught with legal and political tensions surrounding copyright law. Consequently, a large portion of the collection containing in-copyright works remains inaccessible for direct reading. This forces researchers to operate within the controlled environment of their research centre’s secure \enquote{data capsule} which not only presents a steeper learning curve but can also inhibit the broader, collaborative evolution of new computational tools and methods that thrive on open data. At the time of writing, the HathiTrust Research Center---which operates this data capsule environment---is scheduled to lose its funding at the end of 2026, with no confirmed successor for computational access to in-copyright materials. These combined issues of representational bias, data quality, and the constraints of a closed or even closing research environment highlight the challenges of using even the largest institutional corpora to study culture.

These limitations have given rise to a new frontier for corpus-based research: the large-scale \enquote{shadow libraries} that operate outside of formal institutions. Platforms like Library Genesis (LibGen), Sci-Hub, and Z-Library have emerged as a direct response to the access problem, aggregating tens of millions of books and articles in defiance of paywalls and copyright restrictions. These are not merely illicit collections; they are vast, user-driven archives whose very existence represents a form of crowd-sourced cultural curation, making them an essential new source of data for the computational social sciences and humanities~\cite{karaganis2018shadow}. The recent development of Large Language Models (\textsc{llm}s) has opened a Pandora’s box in this regard, moving the use of these datasets from a niche practice to a central component of modern \textsc{ai}. Major companies have publicly acknowledged using such sources in academic papers~\cite{lu2024deepseek} and court cases~\cite{KadreyMeta2025,Anthropic2025}, where their legal defence of \enquote{fair use} has already been partially upheld.

The rationale for our study derives from this context. While these new data sources have been processed as bulk, undifferentiated training data, their use has yet to permeate \textsc{css} research. The poor quality of shadow libraries’ metadata hinders the precise identification and sampling of content. Such precision is essential for representing a period, a style, or literary culture as a whole, and for navigating between these abstractions and their precise instances---that is, clearly identified books.

To bridge this gap, this paper introduces \emph{MajinBook}, an open catalogue designed to resolve this metadata challenge and unlock the full potential of these vast literary archives for cultural analytics. Our method leverages metadata from both shadow libraries and a social reading platform to construct a cohesive bibliographic scaffold. This scaffold binds available editions to their original work and first publication date, resulting in a high-precision corpus of over half-a-million books spanning three centuries, along with secondary datasets to foster future works.

This paper is organised as follows. First, we introduce our data sources and their initial processing, namely the shadow libraries LibGen and Z-Library, and the social network Goodreads. We then detail our linkage strategy, its evaluation and outcome. Finally, we conclude our paper with some legal considerations about our endeavour.

\section*{Shadow Libraries}
\label{sec:shadow_libs}

\subsection*{Library Genesis, Z-Library and Anna's Archive}

The shadow libraries that form the basis of our corpus have distinct but overlapping histories. The oldest and most foundational is Library Genesis (LibGen)\footnote{Throughout this paper, \enquote{LibGen} refers to the original \texttt{libgen.rs} project, rather than any of the numerous forks established since its inception. We do not provide \textsc{url}s for several platforms mentioned in this paper due to the ephemeral nature of their domain names.}, which was started around 2008 to consolidate and share academic texts. Its origins are rooted in the clandestine \enquote{samizdat} culture of the Soviet era and an ethos of providing free access to knowledge for academic communities facing economic hardship and institutional collapse. A pivotal moment in its history occurred in 2011, when LibGen absorbed the massive collection of the defunct shadow library \texttt{Library.nu}, transforming it from a primarily Russian-language archive into a global, multidisciplinary resource.

LibGen’s core operational principle is to be radically open, distributing not just its content but also its catalogue and source code to allow for a resilient, mirrored ecosystem. Appearing around the same time, Z-Library launched in 2009 and grew into one of the largest shadow libraries, with a collection that partially overlaps with LibGen’s but is separately administered and arguably less open~\cite{karaganis2018shadow}. Both platforms have faced significant legal challenges from publishers, culminating in Z-Library having several of its domains seized by the \textsc{fbi} in late 2022.

The most recent development in this ecosystem is Anna’s Archive, which appeared in 2022 and aims to provide a comprehensive, searchable index and mirror of other shadow libraries, including LibGen and Z-Library. For this study, we used LibGen’s own platform to access their data but relied on Anna’s Archive to source Z-Library’s content.

Combining the metadata of Z-Library as of January 2025 and Libgen as of March 2025 yields a list of $77,567,282$ items, of which the vast majority are declared as \textsc{pdf} ($77.9\%$) while $19.2\%$ are referenced as \textsc{epub}\footnote{We classify as \textsc{epub} any extension easily convertible to this open format without significant information loss, namely \texttt{.mobi}, \texttt{.azw(3)} and \texttt{.fb2}}. The remaining $3.5\%$ comprise various formats spanning from raw text files (\texttt{.txt}, \texttt{.rtf}) to word processing extensions (\texttt{.doc}, \texttt{.odf}), along with more exotic types given the context, such as \texttt{.exe} or \texttt{.iso}---all of which were discarded.

\begin{figure*}[t]
  \centering
  \subfloat[Shadow Libraries\label{fig:time_sl}]{
    \includegraphics[width=.32\textwidth]{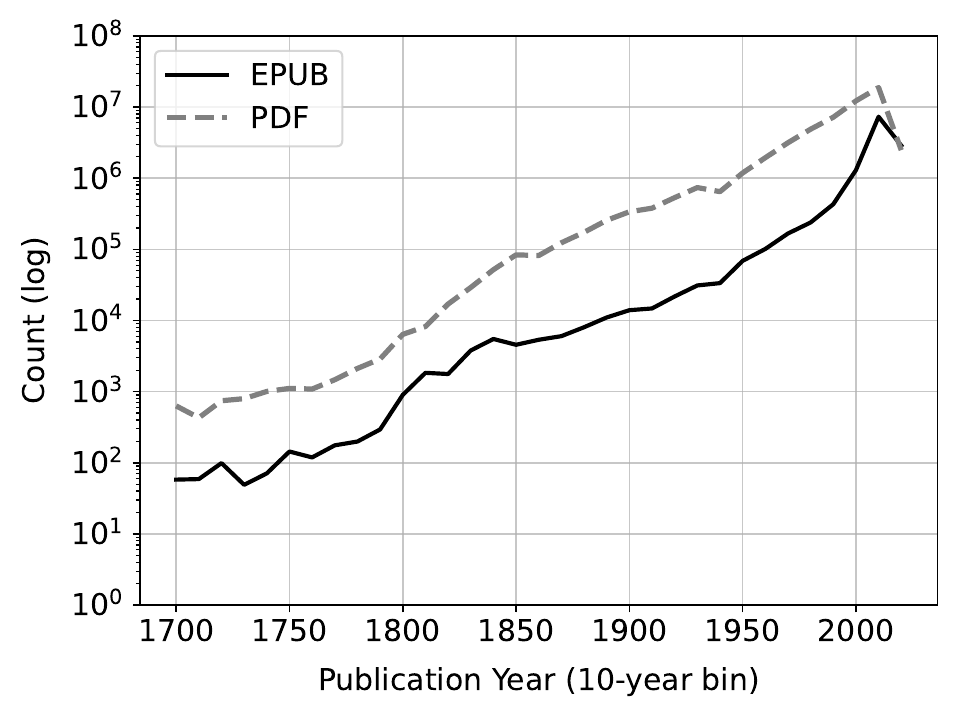}
  }\hfill
  \subfloat[HathiTrust \& Goodreads (editions)\label{fig:time_hathi}]{
    \includegraphics[width=.32\textwidth]{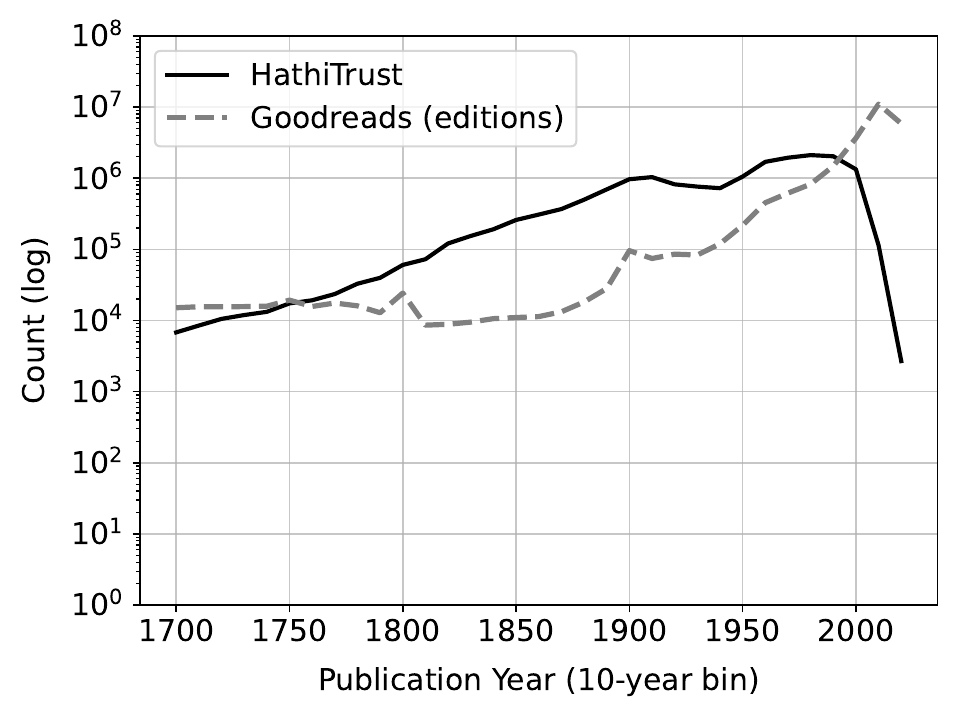}
  }\hfill
  \subfloat[MajinBook \& Goodreads (works)\label{fig:time_majin}]{
    \includegraphics[width=.32\textwidth]{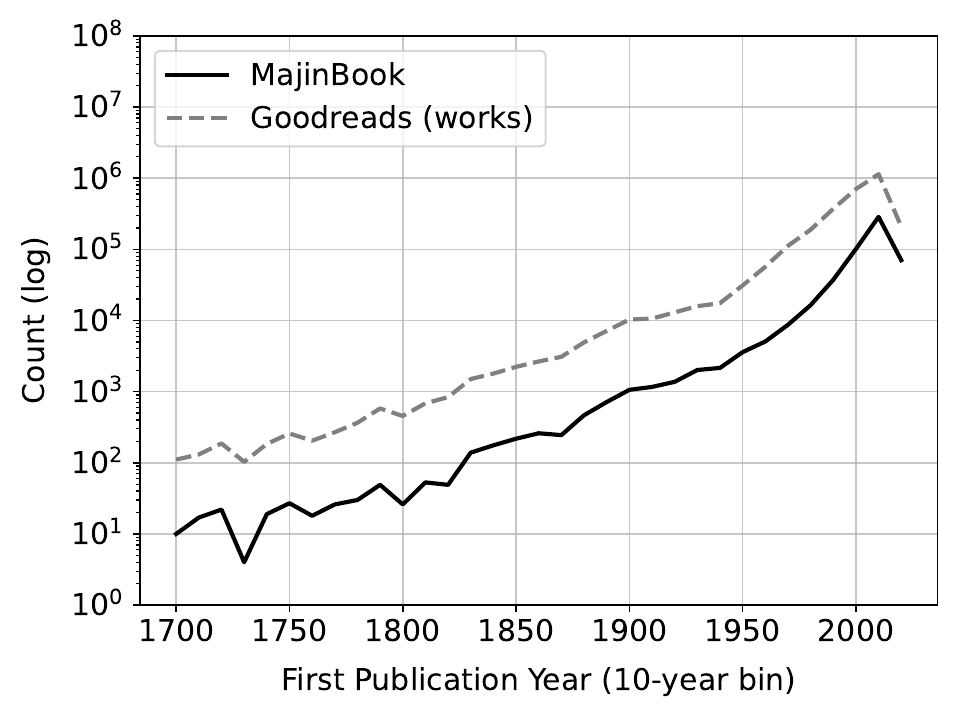}
  }
  \caption{\centering{\textbf{Temporal distributions} and biases of key corpora.\\
  \parbox[t]{\linewidth}{\centering\footnotesize\itshape\setlength{\baselineskip}{1.05\baselineskip}
The figure illustrates the distinct temporal biases of the key corpora, justifying our methodological focus on natively digital content.
    All three plots are semi-logarithmic (log y-axis), displaying item counts binned by publication decade.
    \textbf{(a)} Compares the \textsc{epub} and \textsc{pdf} subsets of shadow libraries.
    \textbf{(b)} Contrasts the scanned HathiTrust corpus with all Goodreads editions.
    \textbf{(c)} Compares our final MajinBook primary corpus (English) to its Goodreads \emph{works} scaffold.
    The plots reveal a fundamental difference in corpus structure. The scanned corpora (\textsc{pdf}, HathiTrust) show relatively stable exponential growth (a linear shape), while the social and natively digital corpora (\textsc{epub}, Goodreads, MajinBook) exhibit super-exponential growth (a convex shape) accelerating towards the present. This validates our decision to discard \textsc{pdf}s and confirms that MajinBook (c) is a representative temporal sample of its source.
  }}}
  \label{fig:timeseries-triptych}
\end{figure*}

\subsection*{Discarding \textsc{pdf}s}
\label{ssec:discarding_pdfs}

The \textsc{pdf} format is highly varied, comprising everything from partial amateur scans to well-indexed official publisher versions. Consequently, consistently parsing this content to extract clean, raw, integral text remains a significant technical challenge. Notable pitfalls include \textsc{ocr} discrepancies and the difficulty of preserving the original content’s reading order. E-books, on the other hand, are natively digital structures---much like web pages---making their content and metadata entirely machine-readable. To include \textsc{pdf} content would be to place items of potentially dubious quality on a par with the very issues found in other scanned corpora, such as HathiTrust, thereby undermining our core objective of a high-quality, \textsc{css}-oriented and natively digital catalogue.

For these reasons, we made the methodological decision to discard all \textsc{pdf}s. This decision introduces a significant temporal bias, favouring recent publications and older works deemed commercially viable enough to be re-issued in a modern digital format. Fig.~\ref{fig:time_sl} clearly illustrates this. On this semi-logarithmic plot, the \textsc{pdf} distribution’s growth is relatively linear, indicating a stable exponential increase over time. In sharp contrast, the \textsc{epub} subset exhibits a convex shape, signalling a super-exponential growth rate that accelerates towards the present.

We argue, however, that this skew is not a simple limitation but a deliberate methodological filter. Rather than a bias against the full shadow library corpus, our choice acts as a \enquote{productive sieve} for a specific, natively digital representation of culture. The trade-off is explicit: we sacrifice the historical completeness of scanned corpora for a dataset of cleaner, more structured, and machine-readable content. Moreover, this productive sieve defines our object of study. We are not claiming to represent \enquote{world literature} in its historical entirety, but rather the digitally mediated canon---culture as it is curated, circulated, and consumed in the 21st century. This temporal bias towards natively digital and commercially viable re-issues is not necessarily a flaw, but a defining characteristic of our corpus.

\begin{table*}[!h]
\vspace{1cm}
\centering
\rowcolors{6}{black!10}{white}
\begin{tabular}{lcccc}
\toprule
& \multicolumn{2}{c}{Shadow Libraries} & HathiTrust & Goodreads \\
\arrayrulecolor{black!45}
\cmidrule(lr){2-3}
\cmidrule(lr){5-5}
\arrayrulecolor{black}
 & \emph{EPUB} & \emph{PDF} &  & \emph{Editions} \\
\midrule
\makecell{N. Items\\\emph{(in millions)}} & 15.2 & 50.5 & 18.9 & 24.2 \\
\arrayrulecolor{black!45}\midrule\arrayrulecolor{black}
\makecell{Herfindahl Index\\\emph{(normalized)}} & 0.24 & 0.74 & 0.32 & 0.57 \\
\midrule
\rowcolor{black!8}English & 47.93 & \textbf{85.97} & 54.95 & 75.44 \\
\rowcolor{white}French & \textbf{8.26} & 1.09 & 7.04 & 4.63 \\
\rowcolor{black!8}German & 5.94 & 3.96 & \textbf{8.61} & 3.66 \\
\rowcolor{white}Spanish & \textbf{9.45} & 0.94 & 5.44 & 3.84 \\
\rowcolor{black!8}Russian & \textbf{4.31} & 3.45 & 2.85 & 0.65 \\
\rowcolor{white}Chinese & \textbf{9.59} & 2.35 & 3.26 & 0.58 \\
\rowcolor{black!8}Italian & \textbf{4.24} & 0.51 & 2.27 & 2.17 \\
\rowcolor{white}Portuguese & \textbf{1.68} & 0.33 & 1.23 & 1.16 \\
\rowcolor{black!8}Dutch & \textbf{2.06} & 0.14 & 0.71 & 0.91 \\
\rowcolor{white}Japanese & 0.75 & 0.09 & \textbf{3.07} & 0.85 \\
\rowcolor{black!8}Polish & \textbf{1.13} & 0.08 & 0.64 & 0.61 \\
\rowcolor{white}Arabic & 0.57 & 0.05 & \textbf{1.05} & 0.45 \\
\rowcolor{black!8}Czech & \textbf{0.39} & 0.03 & 0.36 & 0.28 \\
\rowcolor{white}Swedish & 0.25 & 0.01 & \textbf{0.50} & 0.38 \\
\rowcolor{black!8}Danish & 0.26 &  & \textbf{0.36} & 0.27 \\
\rowcolor{white}Hungarian & \textbf{0.49} & 0.10 & 0.26 & 0.17 \\
\rowcolor{black!8}Korean & 0.29 & 0.02 & \textbf{0.36} & 0.09 \\
\rowcolor{white}Turkish & 0.14 & 0.12 & 0.21 & \textbf{0.52} \\
\rowcolor{black!8}Bulgarian & \textbf{1.36} & 0.11 & 0.14 & 0.17 \\
\rowcolor{white}Indonesian &  & 0.10 & \textbf{0.28} & 0.18 \\
\rowcolor{black!8}Romanian & 0.14 & 0.04 & 0.13 & \textbf{0.29} \\
\rowcolor{white}Ukrainian & 0.06 & 0.11 & \textbf{0.15} & 0.09 \\
\rowcolor{black!8}Persian &  & 0.02 & 0.17 & \textbf{0.19} \\
\rowcolor{white}Greek & 0.01 & 0.07 & 0.12 & \textbf{0.23} \\
\rowcolor{black!8}Catalan & \textbf{0.15} &  & 0.07 & 0.11 \\
\rowcolor{white}Serbian & 0.02 & 0.01 & 0.16 & \textbf{0.17} \\
\rowcolor{black!8}Norwegian &  & 0.01 & \textbf{0.21} & 0.14 \\
\rowcolor{white}Hebrew & 0.07 & 0.02 & \textbf{0.46} & 0.08 \\
\rowcolor{black!8}Lithuanian & \textbf{0.10} & 0.04 & 0.02 & 0.09 \\
\rowcolor{white}Bengali & 0.05 & 0.06 & \textbf{0.11} & 0.07 \\
\rowcolor{black!8}Finnish & 0.02 &  & 0.10 & \textbf{0.29} \\
\rowcolor{white}Croatian & 0.01 & 0.01 & \textbf{0.23} & 0.11 \\
\rowcolor{black!8}Vietnamese & 0.02 & 0.01 & \textbf{0.11} & 0.09 \\
\rowcolor{white}Slovak & 0.04 &  & 0.06 & \textbf{0.09} \\
\rowcolor{black!8}Thai &  & 0.01 & \textbf{0.19} & 0.09 \\
\rowcolor{white}Latin &  & 0.02 & \textbf{0.83} & 0.07 \\
\rowcolor{black!8}Hindi & 0.02 & 0.01 & \textbf{0.25} & 0.07 \\
\rowcolor{white}Afrikaans & \textbf{0.05} & 0.01 & 0.03 & \textbf{0.05} \\
\rowcolor{black!8}Latvian & \textbf{0.05} & 0.01 & 0.02 & \textbf{0.05} \\
\rowcolor{white}Slovenian &  &  & \textbf{0.07} & \textbf{0.07} \\
\bottomrule
\end{tabular}

\caption{\centering{\textbf{Comparative Analysis of Corpus Scale and Linguistic Diversity.}\\
\parbox[t]{\linewidth}{\centering\footnotesize\itshape\setlength{\baselineskip}{1.05\baselineskip}
    This table provides the core quantitative justification for our methodological decision to focus on the \textsc{epub} subset. 
    It compares the scale (in millions of items) and linguistic concentration (normalised Herfindahl Index) of the \textsc{epub} and \textsc{pdf} shadow library subsets against the HathiTrust and Goodreads corpora. The analysis reveals a stark trade-off: the \textsc{pdf} corpus, despite its size, is linguistically homogeneous (HI=$0.74$), with English comprising $85.97\%$ of its content. In contrast, the \textsc{epub} subset (our chosen base) is the most linguistically diversified of all corpora (HI=$0.24$), significantly outperforming even the HathiTrust collection (HI=$0.32$). This demonstrates that our filtering process, while reducing the total item count, produced a smaller but far more balanced and representative dataset for cultural analytics.
}}}
\label{tab:langs}
\end{table*}

This contemporary constraint yields a significant and somewhat counter-intuitive benefit. As Table~\ref{tab:langs} shows, the \textsc{epub} subset is far more linguistically diversified (Herfindahl Index (\textsc{hi})~$= 0.24$) than its \textsc{pdf} counterpart (\textsc{hi}=$0.74$). Notably, it is also less concentrated than the HathiTrust corpus (\textsc{hi}=$0.32$), demonstrating that our \enquote{sieve} produces a dataset that is both high-quality and linguistically diverse. This resulting language fragmentation---which reduces the share of English from over $85$\% in the \textsc{pdf} set to just $48$\% in our corpus (Table~\ref{tab:langs})---can be seen as a positive indicator of cultural breadth, enhancing the dataset's representativeness for cultural analytics.

\section*{Goodreads}
\label{sec:goodreads}

Goodreads~\cite{goodreads2006www}, a prominent social reading platform launched in 2006 and acquired by Amazon in 2013, boasts over 150 million members who rate, review, catalogue, and discuss books, creating a vast digital archive of contemporary reader responses and amateur criticism. Academics utilise this extensive dataset to analyse reading activity at scale, compare modern literary reception with historical patterns, and understand how works gain or lose popularity~\cite{walsh2021goodreads,antoniak2024afterlives,bourrier2020social,kousha2017goodreads,hu2025decides}. Nevertheless, these studies acknowledge limitations, including demographic biases within the user base (predominantly white, female, and \textsc{us}-centric) and the potential for review manipulation.

We first detail our data requirements and the rationale for our choice of source, before describing our crawling methodology and its outcome.

\subsection*{Data Requirements and Rationale}
\label{subsec:data_requirements_and_rationale}

\subsubsection*{Bibliographic Scaffold}

Homer’s \emph{Odyssey} can be understood as an abstract cultural artefact; a shared reference to ancient history, cunning, and homecoming. However, each physical copy of the text is a unique object that bears witness to the technologies of its creation, the history of its translations, and the pressures of censorship. The word \emph{book} often carries both of these meanings. To be more precise, we use the term \emph{work} to refer to the abstract text---the cultural artefact---and \emph{edition} to refer to a specific published version. To some extent, the edition is the object and the work is the idea. To study the latter, one must necessarily go through the former---computationally or otherwise.

This Work-Edition scaffold partly relates to more advanced classification systems, for instance the \emph{Functional Requirements for Bibliographic Records} (\textsc{frbr})~\cite{funcReq1998}, devised by the International Federation of Library Associations and Institutions (\textsc{ifla}). Their \emph{Work} entity aligns closely with our eponymous category, both referring to a \enquote{distinct intellectual or artistic creation}. With a degree of interpretative flexibility---and in the specific context of books---our notion of edition corresponds roughly to the grouping of \textsc{frbr}’s Expression and Manifestation entities. This approach also conceptually parallels the method used by the Online Computer Library Center (\textsc{oclc}) to cluster bibliographic records into \emph{Work} entities within WorldCat.

A key advantage of this system is that it binds each \emph{edition} not to its own publication date, but to the \emph{first} publication date of its corresponding work. This temporal grounding is crucial for diachronic analysis, as it anchors the work as a cultural representation to the period in which it was primarily written. This prioritises the history of literary production over the history of reception. While historians of the book might prefer to track the circulation of specific editions, our approach is designed to map, and therefore focus on, the cultural onset of the intellectual artefact. Simultaneously, the inclusion of Goodreads metadata offers a rare quantitative proxy for modern reception---capturing how these historical works are read and valued by users today.

In practice, we adopt the internal \emph{work-edition} mappings provided by Goodreads. These mappings form the core structure of its platform and proved to have very few inconsistencies. In the dataset described below, $60.9\%$ of the works feature a precise first publication date.

\subsubsection*{Genres and Popularity Metrics}

While the first publication date is crucial for \textsc{css} researchers to select works from a specific period, other partitions in the data can prove highly relevant to narrow the corpus based on alternative criteria. First and foremost, in the context of books, \emph{genres} are a common classification enabling thematic differentiation. The precise internal classification mechanism for genres on Goodreads remains opaque, but the \enquote{genres} section associated with most books is established at the work level, i.e., all editions of a given work share the same genres. These categories appear to be the product of a top-down curation of the platform’s crowd-sourced \enquote{shelves} that enable users to organise their books. Certain shelves, such as \enquote{fiction} or \enquote{romance}, are promoted to the \enquote{genres} section, whereas others, such as \enquote{i-own} or \enquote{to-read}, are excluded.

Popularity is another common corpus selection criterion. Indeed, the very inclusion of a work in any corpus \emph{represents} a form of selection against obscurity, an escape from what Moretti termed \emph{The Slaughterhouse of Literature}~\cite{moretti2000slaughterhouse}. Different corpora embody distinct curation models: for instance, the HathiTrust dataset, composed of contributions from partner libraries, represents a decentralised yet expert-driven curation model, while shadow libraries are largely crowd-sourced. Regardless of the base corpus, distinct popularity metrics can prove useful to differentiate between popular and obscure works. Goodreads, as a social network, offers typical features such as ratings and reviews. Like HathiTrust and shadow libraries, these metrics embody a specific survivorship bias that characterises the form of popularity they convey. In that regard, Goodreads ratings reflect approximately two decades of user engagement (2006--present) and should therefore be understood as a snapshot of contemporary reading preferences rather than a stable historical measure of popularity. Integrating these features into our catalogue offers significant analytical potential, allowing researchers both to study platform-specific popularity patterns and to select sub-corpora based on specific popularity criteria. This latter point is particularly relevant for computationally intensive \textsc{ai} research, where budget is a significant constraint. Quantitative popularity metrics allow a researcher to create an affordably-sized sub-corpus by sampling the most visible works, while still preserving the catalogue’s thematic and temporal representativeness---subject to the biases acknowledged above.

\subsubsection*{Goodreads v.\ Other Data Sources}

Before selecting Goodreads, we explored several alternative data sources. We believe it is relevant for future research to make our rationale for this choice explicit. The most significant alternative was Open Library~\cite{openlibrary2006www}, a non-profit initiative of the Internet Archive, launched in 2006 with the ambitious goal of creating \enquote{one web page for every book ever published}.

On the surface, Open Library is a compelling choice. Its primary strength is its open nature, providing free, bulk access to one of the largest structured bibliographic datasets in the world. It boasts tens of millions of records, covers a rich long-tail of lesser-represented languages, and utilises a formal work-edition data model. However, the project's greatest strength---its open, wiki-based contribution model---is also its most significant weakness. The metadata is unreliable, containing frequent duplicate records and incomplete entries. More critically for our study, we identified two fatal flaws: a low coverage rate for \emph{first publication date} and pervasive inconsistencies in the work-edition linkages.

We also evaluated Wikipedia as a potential alternative or complementary source. Its appeal was twofold. First, it offers a distinct form of curation: a book's existence as a dedicated Wikipedia article signals a type of \enquote{encyclopaedic notability} that is different from, and complementary to, the social popularity metrics of Goodreads. Second, its underlying data model (via Wikidata) is often structured around \emph{works} and \emph{first publication dates}, aligning well with our bibliographic scaffold.

However, we encountered a significant technical challenge in isolating the correct entities. While many books are instances of the \enquote{written work} category, this exists within a deep and complex hierarchical system. It proved difficult to reliably discriminate actual books from other types of written works, such as articles or pamphlets. Given this difficulty in reliably scoping the corpus, we set this promising source aside for the current study.

We therefore accept this trade-off. In exchange for the high-quality, consistent metadata that Open Library lacks, and in avoidance of the entity-scoping ambiguity from Wikipedia we could not resolve, we build our scaffold upon Goodreads. We explicitly acknowledge this foundation is not neutral, but is a structure that reflects a social and commercial curation of literature. Our dataset is therefore not a representation of \enquote{world literature} in its entirety, but a representation of 21st-century literary culture as it is mediated by a major commercial platform.

\subsection*{The Crawl of Goodreads}
\label{subsec:crawl_goodreads}

To harvest data from the Goodreads website\footnote{\texttt{goodreads.com}}, we gathered $1{,}225{,}390$ editions to serve as seeds. These were drawn from the platform’s various public book aggregations: $30$ tags, $642$ shelves, $1{,}419$ genres, $8{,}194$ awards, and $30{,}497$ lists. Together, these seed editions corresponded to $990{,}010$ distinct works by $927{,}707$ authors. This initial set of works comprised a total of $11{,}319{,}891$ editions.

We then recursively expanded this baseline by collecting works from the platform’s recommendations (such as those featured in the \enquote{Readers also enjoyed} section), as well as other relevant works by the authors already gathered. To mitigate the long tail of obscure items, we applied a specific inclusion threshold for an author’s additional works: they were required to have at least one rating and two editions.

This iterative crawl continued until the author-led discovery of new items plateaued at depth four and ceased at depth five, coinciding with the complete exhaustion of the recommendation-led crawl. These simultaneous terminations suggest that our dataset is a comprehensive representation of Goodreads' discoverable content.

This approach effectively neutralises the risk of algorithmic selection bias. By exhausting the network, we eliminate the path-dependency of the recommender system, yielding a corpus defined by the platform’s boundaries rather than its suggestions.

\begin{figure}
    \centering
    \includegraphics[width=\columnwidth]{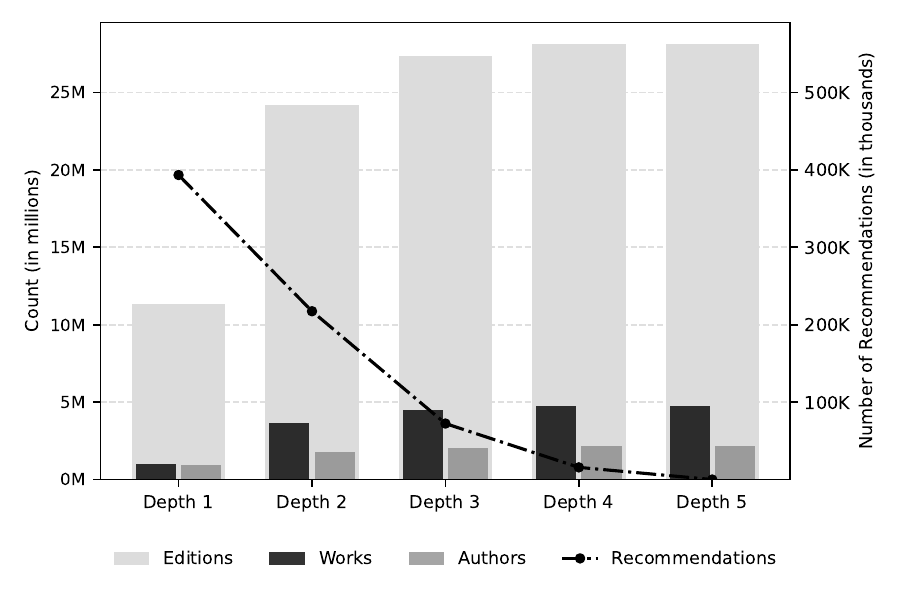}
    \caption{\textbf{The crawl of Goodreads}: Item acquisition and recommendation decay.\\
    \parbox[t]{\linewidth}{\footnotesize\itshape\setlength{\baselineskip}{1.05\baselineskip}%
      The figure illustrates the efficiency of our crawl methodology. The bars show
      the cumulative counts of Editions, Works, and Authors (left axis, in millions)
      gathered at each stage. The line plot tracks the number of new Recommendations
      (right axis, in thousands) discovered at each depth. The plot reveals a power-law
      distribution: the initial depths rapidly capture the most prominent items, while
      subsequent depths explore a long tail of less-connected content.%
    }}
    \label{fig:goodreads}
\end{figure}

The final dataset comprises $4{,}778{,}124$ works, $28{,}105{,}913$ editions, and $2{,}150{,}522$ authors (cf. Fig.~\ref{fig:goodreads}). In total, $15$\% of works were discovered via recommendations, $20$\% came from the seed set, and $65$\% were retrieved through author-based expansion.

To the best of our knowledge, and based on our review of the state of the art~\cite{DBLP:conf/recsys/WanM18,DBLP:conf/acl/WanMNM19,JannesarGoodreads2020,BrightDataGoodreads2024,hu2025decides}, this represents one of the largest publicly available crawls of Goodreads data published to date.

\section*{MajinBook}
\label{sec:majinbook}

\subsection*{Preparing Data for Matching}

Our matching strategy relied on \emph{book identifiers}, primarily the \emph{International Standard Book Number} (\textsc{isbn}) and the \emph{Amazon Standard Identification Number} (\textsc{asin}), extracted from both shadow library metadata and the \textsc{epub} files themselves. We observed that these identifiers were often unreliable for matching a file to its \emph{exact} edition. However, we found that even an 'incorrect' identifier was still a highly reliable pointer to the correct \emph{work}. We therefore designed our strategy around this insight, linking files to a robust \texttt{work-language} pair rather than pursuing a more fragile edition-level match.

From the \textsc{epub} subset of the shadow libraries catalogue, we harvested all downloadable files. These were then standardised to a consistent \texttt{.epub} format and converted into raw, un-marked-up text files; any error during this process resulted in the file being discarded. This initial filtering yielded our base dataset of $11{,}130{,}032$ files. We then applied a size filter, discarding $213{,}167$ items deemed too small ($<10$\textsc{KB}, ca.\ 4 \textsc{A4} pages) or too large ($>10$\textsc{MB}, ca.\ $4{,}000$ \textsc{A4} pages) to be actual books.

First, we computed a $128$-element \emph{MinHash} signature for every item. We then used \emph{Locality-Sensitive Hashing} (\textsc{lsh}) to efficiently identify, for each item, all items with an approximate Jaccard similarity of $0.8$ or higher.\footnote{We used the Python library \emph{datasketch} which automatically computed the optimal parameters of $9$ bands and $13$ rows to find pairs at this similarity level (\texttt{github.com/ekzhu/datasketch}).} This $0.8$ threshold is a conservative standard for identifying near-duplicates, while still allowing for minor textual variations between different editions.

This enabled us to group items as potential duplicates or different editions from the same work in a given language, yielding a list of $1{,}954{,}010$ unique clusters with more than one item, $77.6\%$ of which had at least one element with at least one \emph{book identifier}. This set of $1{,}516{,}332$ clusters of identifiable shadow library items constitutes our base for matching with the Goodreads metadata.

To construct our Goodreads matching set, we filtered our $4.78$ million works for those that included a \emph{first publication year}, an essential field for our diachronic scaffold. This step yielded our base dataset of $2{,}904{,}994$ works ($60.9\%$). This filtered set is highly suitable for the matching task, as $99.7\%$ of these works also feature at least one book identifier.

This significant reduction is a deliberate methodological choice, not a simple loss of data. We found strong evidence that the $39.1\%$ of works we discarded are, on average, less central to the platform and of lower data quality. For instance, works without a \emph{first publication year} are significantly less evaluated, with a median number of ratings of $9$ (\textsc{iqr}$=2$--$41$), compared to $17$ (\textsc{iqr}$=4$--$99$) for the works we retained. This suggests they are less engaged with by the platform's users. Furthermore, this metadata-completeness correlates with discoverability: a \emph{first publication year} was present on over $80\%$ of items in the initial crawl depths but dropped progressively to the final $60\%$ average. This indicates that the most discoverable items on Goodreads are also the most likely to have complete metadata.

\subsection*{Matching and Evaluation}
\label{sec:match_eval}

Our strategy for matching shadow library \emph{clusters} with Goodreads \emph{works} relied on identifier overlap. A given cluster was linked to a given work if the set of all identifiers found within that cluster's items had a non-empty intersection with the set of all identifiers aggregated from that work's editions. Any such match flagged the whole cluster as a potential \emph{candidate} for that work, tagged with the cluster's language. This process enabled us to link $770{,}840$ Goodreads works to at least one cluster. In total, these retained clusters comprise $4{,}216{,}400$ shadow library items.

Deeming identifier-based links to be a necessary but not sufficient condition, we implemented a second verification step using metadata. This step performs an exhaustive title comparison for each candidate. For a given cluster and a potential work, we compared every title within the shadow library cluster against every edition title for that work (in the cluster's language). For each of these pairs, we computed a partial ratio fuzzy match, yielding a score from $0$ to $100$. Finally, we calculated the mean of this entire comparison matrix to produce a single, robust \emph{title score} for the candidate. Overall, the distribution of title scores is highly left-skewed, with a median score of $99.1$ (\textsc{iqr}$=86.6$--$100.0$). This indicates that the vast majority of identifier-based matches are confirmed by our title matching method.

We then conducted a human evaluation study to validate our matching method and determine an optimal \emph{title score} threshold. For this, we sampled 200 English-language candidates, using a stratified approach that deliberately overrepresented items with lower, more ambiguous scores.\footnote{More precisely, we selected $5$ items with a \emph{title score} between $[0,20]$, $15$ in $]20,40]$, $10$ in $]40,50]$, $15$ in $]50,60]$, $25$ in $]60,70]$, $30$ in $]70,80]$, $50$ in $]80,90]$, and $50$ in $]90,100]$.} We built a simple web interface for this task, which, for each candidate, displayed the corresponding Goodreads title and authors alongside the first 100 paragraphs of a random item from the shadow library cluster.

\begin{figure}[h!]
    \centering
    \includegraphics[width=\columnwidth]{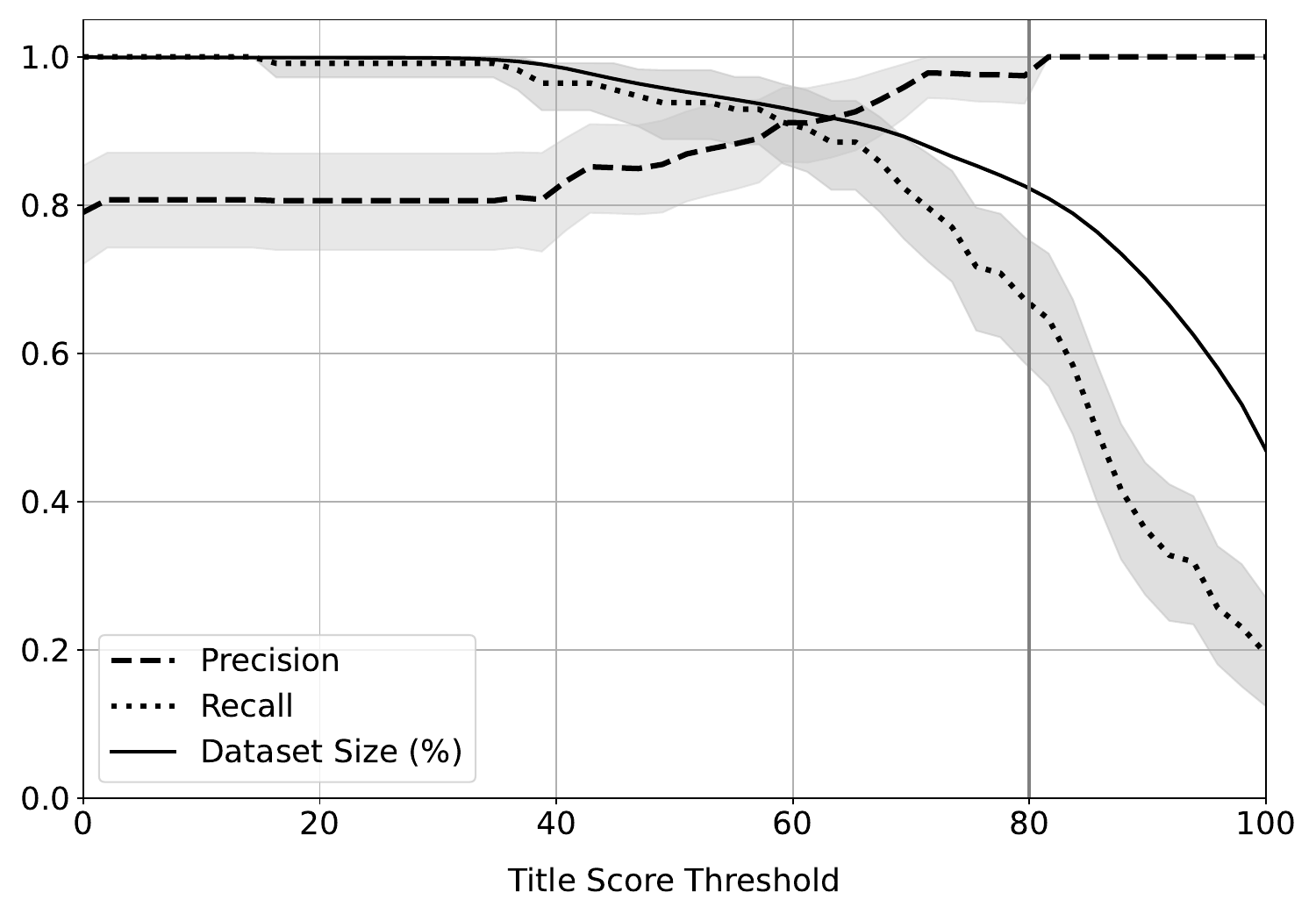}
    \caption{\textbf{Precision-recall trade-off for book matching} based on the title score threshold.\\
    \parbox[t]{\linewidth}{\footnotesize\itshape\setlength{\baselineskip}{1.05\baselineskip}%
        The plot shows the point estimates for precision (dashed line) and recall (dotted line), along with their 95\% confidence intervals (shaded areas), derived from bootstrap resampling of 143 human evaluations. 
        The solid black line indicates the percentage of the dataset retained at each threshold. 
        A vertical line marks our chosen operational threshold of 80, which prioritises high precision for the final catalogue.
    }}
    \label{fig:human_eval}
\end{figure}

Evaluators from our lab were then asked to assess each candidate and assign one of three labels: \enquote{Yes} (a correct match), \enquote{No} (an incorrect match), or \enquote{I don't know/I can't tell}. We concluded the evaluation once every item in the sample had received at least one review. Figure~\ref{fig:human_eval} shows the results for the $143$ items for which evaluators could come to a conclusion and plots the resulting precision, recall, and dataset size as a function of the \emph{title score} threshold. The figure clearly illustrates the classic trade-off: as the threshold increases, precision rises, while recall and dataset size fall. Based on our stated goal of a high-precision catalogue, and supported by this evaluation, we selected a final threshold of $80$, which achieves a precision of nearly $1.0$ (Fig.~\ref{fig:human_eval}).

The 57 items ($28.5\%$) that evaluators labelled \enquote{I don't know/I can't tell} were not factored into the primary precision-recall computation. These items were not determined to be incorrect, but were rather \emph{un-evaluatable} by our lab colleagues, as they typically lacked title pages and started \emph{in media res}, offering few visual cues for confirmation. To determine whether a high \emph{title score} could be trusted on these ambiguous files, we conducted a deeper, secondary analysis on the $24$ ambiguous items with a title score above our operational threshold ($80$) and concluded that $22$ were correct matches ($91.7\%$ precision). The $2$ errors were not random: one was a single book matched to a $3$-book set, and the other was a different book by the right author. This demonstrates that the \emph{title score}'s predictive power is robust even for files that are visually ambiguous. It confirms that the formatting issue is statistically independent of the score's relevance and validates our decision to proceed with the threshold derived from the $143$ conclusively-labelled items.

\subsection*{Release}
\subsubsection*{Primary Dataset: the MajinBook Catalogue}

We only evaluated our matching methodology on English titles, and $82.9\%$ of our retained matches are for English-language content. This English subset, numbering $539{,}530$ items, therefore constitutes our primary dataset, fulfilling the large-scale, high-quality ambition of our study.

Each entry in this final catalogue includes the following metadata fields:

\begin{itemize}[noitemsep, topsep=0pt]
    \item Goodreads Work \textsc{id}
    \item First publication year
    \item Authors' full names and Goodreads \textsc{id}s
    \item Title
    \item Rating and number of ratings
    \item Shadow Libraries \textsc{id}s corresponding to this Work in English
\end{itemize}

Additionally, $84.04\%$ of the catalogue features a list of \emph{Genres} and $99.23\%$ includes the number of reviews.

\subsubsection*{Secondary Datasets}

\begin{figure}
    \centering
    \includegraphics[width=\columnwidth]{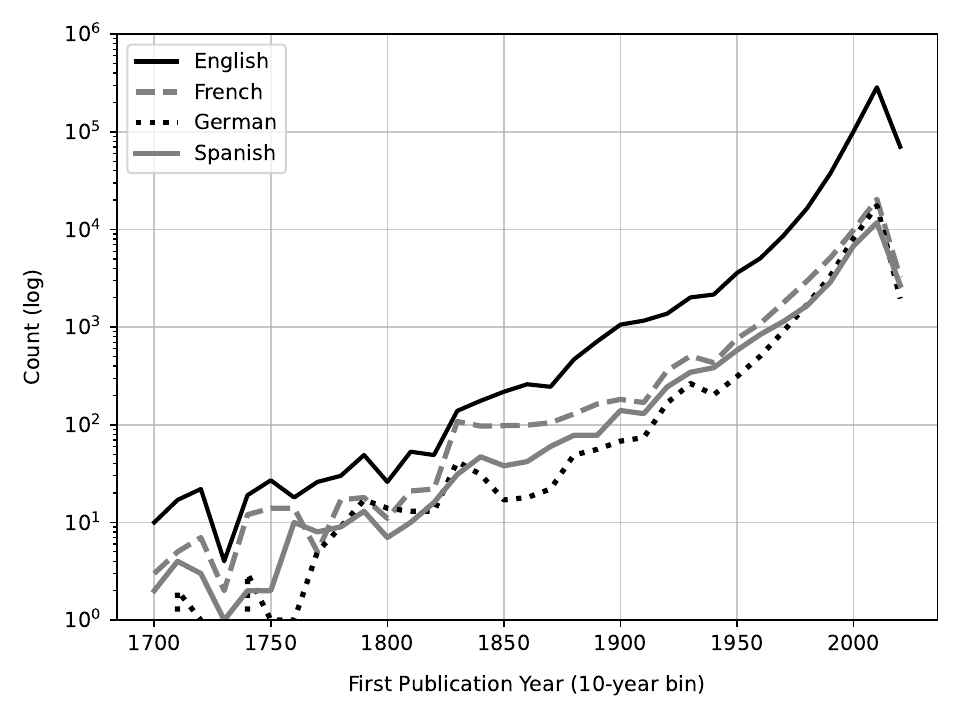}
    \caption{\textbf{Temporal distribution} of primary (English) v. secondary datasets.}
    \label{fig:majin_all_langs}
\end{figure}

Despite the overwhelming dominance of English, our methodology captured significant volumes of content in other languages with robust temporal distributions (Fig.~\ref{fig:majin_all_langs}). Although we did not conduct human evaluation for these languages, their \emph{title score} distributions are highly left-skewed, closely mirroring the distribution of our validated English set.

While this similarity is a promising heuristic, it is not a substitute for formal validation. Based on these two criteria—significant volume and a comparable title score distribution—we selected three of the largest non-English corpora for release: namely French ($47{,}960$ items), German ($35{,}559$), and Spanish ($30{,}169$). The sharp drop in volume relative to the source shadow libraries is explained by the fact that our dataset is constrained by the linguistic distribution of Goodreads (cf. Table~\ref{tab:langs}).

The feature coverage for these datasets is identical to the primary English corpus, with the sole exception of \emph{Genres}, which falls to $77.4\%$ for Spanish, $75.5\%$ for German, and $70.1\%$ for French. We hypothesise that this is not an artefact of our matching process but rather reflects the sparser metadata available for non-Anglophone works within Goodreads itself.

We must, however, stress that the precise quality of these matches remains unverified. We release these secondary catalogues as experimental datasets to foster future work, noting that they do not carry the same high-precision guarantee as our primary English corpus.

\subsubsection*{Underlying Datasets}

Finally, in addition to the primary catalogues, we are releasing the underlying datasets that enabled this study, namely:
\begin{itemize}[noitemsep, topsep=0pt]
    \item The formatted metadata for the retained elements from the shadow libraries \textsc{epub} subset.
    \item The formatted metadata extracted for every Work, Edition, and Author from the Goodreads crawl.
    \item The MinHash signatures for every item considered in the shadow libraries.
\end{itemize}

\section*{Legal and Ethical Considerations}
\label{sec:legal}

The legality of using shadow libraries for research remains a contested and evolving issue, as illustrated by recent lawsuits involving major \textsc{ai} companies~\cite{KadreyMeta2025,Anthropic2025}. At the time of writing this paper, legal frameworks in many countries are being adjusted to find a balance between copyright law and the \textsc{ai} use of copyrighted materials~\cite{sag2024globalization}. Our project’s legal standing rests on a critical distinction between our final product (a public, metadata-only catalogue) and our process (acquiring and analysing data from Goodreads and shadow libraries).

MajinBook's final output is a bibliographic index. The titles, author names, publication dates, and identifiers that populate our catalogue are factual data, not copyrightable expression~\cite{feist1991}. Moreover, our index reproduces no textual content from the underlying books---a posture considerably more restrained than the snippet display upheld as fair use in \emph{Authors Guild v. Google}~\cite{AuthorsGuildGoogle2015} (2015), where the court permitted the copying of entire books for a transformative search index.

Goodreads data was acquired by scraping publicly accessible pages without circumventing any authentication mechanism or technical protection measure. Following \emph{hiQ Labs v. LinkedIn}~\cite{hiQLabs2022} (2022), such access to publicly available data does not constitute \enquote{unauthorised access} in the sense of the Computer Fraud and Abuse Act (\textsc{cfaa}). From a European standpoint, where the data was acquired, this falls squarely within the Text and Data Mining (\textsc{tdm}) exception for research purposes, which provides a mandatory research exception~\cite{DSMDirective2019} that cannot be overridden by contract (Art. 7). Even if Goodreads' terms of service were to prohibit automated access to their platform, this regulation grants academics acting in good faith and with no commercial purpose the right to do so.

Lastly, harvesting and computing data from shadow libraries as we did can be considered through the lens of several legal frameworks: In the \textsc{eu}, the above-mentioned \enquote{\textsc{tdm} exception}, and in the \textsc{us}, the fair use doctrine incorporated into the Copyright Act of 1976~\cite{CopyrightAct1976} complemented by the recent Text and Data Mining exemption to the Digital Millennium Copyright Act~\cite{DMCA20140}. The precise application of these frameworks is a complex and evolving debate, one that is beyond the scope of this paper and our expertise as non-legal scholars. However, from our reading, the core of the discussion---both legal and ethical---appears to hinge on the intent and context of the use.

Our work is situated in a public research institution, with no commercial interest, and aims to foster public knowledge about language, literature, and cultural representations. Applying the factors commonly weighed by these frameworks to our endeavour: the purpose is non-commercial academic research conducted in good faith; the data was accessed through publicly available channels, though the legal status of the content hosted on shadow libraries remains contested; the amount of information extracted from each book is minimal (metadata, hash signatures), serving a purely bibliographic purpose; and the resulting catalogue neither substitutes for any book we may have accessed nor affects their potential market.

\section*{Data availability}

The datasets generated and/or analysed during the present study are available in the Zenodo public data repository: \href{https://doi.org/10.5281/zenodo.17609566}{doi.org/10.5281/zenodo.17609566}.

The detailed description of the datasets’ structure and metadata schemas is available on GitHub: \href{https://github.com/mazieres/MajinBook}{github.com/mazieres/MajinBook}.

\section*{Acknowledgements}

The authors thank Céline Castets-Renard and Benoît de Courson for their valuable input on this research.

This work was funded in part thanks to the support of \textsc{prairie}-\textsc{psai} (Paris Artificial intelligence Research institute--Paris School of Artificial Intelligence, reference \texttt{ANR-22-CMAS-0007}).

{
\footnotesize
\bibliographystyle{ieeetr}
\bibliography{bibliography}
}

\end{document}